\documentclass[11pt,a4paper]{article}
\pdfoutput=1
\usepackage[hyperref]{naaclhlt2018}
\usepackage{times}
\usepackage{latexsym}

\usepackage{multirow}
\usepackage{comment}
\usepackage{amsmath}
\usepackage{graphicx}
\aclfinalcopy % Uncomment this line for the final submission
\title{Assessing Language Proficiency from Eye Movements in Reading}

\author{Yevgeni Berzak \\
  MIT BCS \\
  {\tt berzak@mit.edu} \\\And
  Boris Katz \\
  MIT CSAIL \\
  {\tt boris@mit.edu} \\\And
  Roger Levy \\
  MIT BCS \\
  {\tt rplevy@mit.edu} \\}

\date{}

\begin{document}

\maketitle

\begin{abstract}
We present a novel approach for determining learners' second language proficiency which utilizes behavioral traces of eye movements during reading. Our approach provides stand-alone eyetracking based English proficiency scores which reflect the extent to which the learner's gaze patterns in reading are similar to those of native English speakers. We show that our scores correlate strongly with standardized English proficiency tests. We also demonstrate that gaze information can be used to accurately predict the outcomes of such tests. Our approach yields the strongest performance when the test taker is presented with a suite of sentences for which we have eyetracking data from other readers. However, it remains effective even using eyetracking with sentences for which eye movement data have not been previously collected. By deriving proficiency as an automatic byproduct of eye movements during ordinary reading, our approach offers a potentially valuable new tool for second language proficiency assessment.  More broadly, our results open the door to future methods for inferring reader characteristics from the behavioral traces of reading.
\end{abstract}

\section{Introduction}

It is currently estimated that over 1.5 billion people are learning English as a Second Language (ESL) worldwide. Their learning progress is commonly evaluated with classroom tests prepared by language instructors, quizzes in language learning software such as Duolingo and Rosetta Stone, and by official standardized language proficiency tests such as TOEFL, IELTS, MET and others. In ``high stakes'' scenarios, official language proficiency tests are the de-facto standards for language assessment; they are accepted by educational and professional institutions, and are taken by millions of language learners every year (for example, in 2016 over three million people took the IELTS test \cite{wiki:ielts}). These tests probe language proficiency based on performance on various linguistic tasks, including grammar and vocabulary exams, reading and listening comprehension questions, as well as essay writing and speaking assignments.

Despite their ubiquity, traditional approaches to language proficiency testing have several drawbacks. First, such tests are typically prepared manually and require extensive resources for test development. Moreover, their validity can be undermined by test specific training, prior knowledge of the evaluation mechanisms \cite{powers2002}, as well as plain cheating via unauthorized access to test materials. Further, the utilized testing and evaluation methodologies vary across different tests, and test materials are in most cases inaccessible to the research community. Perhaps most crucially, the reliance of these tests on the end products of linguistic tasks makes it challenging to study learners' language processing patterns and the difficulties they encounter in real time.

In this work we propose a novel methodology for language proficiency assessment which marks a significant departure from traditional language proficiency tests and addresses many of their drawbacks. In our approach, we determine language proficiency from broad coverage analysis of \emph{eye movements during reading of free-form text in a foreign language}, a special case of the general problem of inferring comprehender characteristics and cognitive state from the measurable traces of real-time language processing. Our framework does not require the test taker to prepare for the test or to perform any hand-crafted linguistic tasks, but simply to attentively read an arbitrary set of sentences. To the best of our knowledge, this work is the first to propose and implement such an approach, yielding a novel language proficiency evaluation scheme which relies solely on ordinary reading.

Our framework builds on previous research in psycholinguistics demonstrating that the eyetracking record reflects how readers interact with the text and how language processing unfolds over time \cite{frazier1982,rayner1998,rayner2012}. In particular, it has been shown that key aspects of the reader's characteristics and cognitive state, such as mind wandering during reading \cite{reichle2010}, dyslexia \cite{rello2015} and native language \cite{berzak2017} can be inferred from their gaze record. Despite these advances, the potential of the rich and highly informative behavioral signal obtainable from human reading for automated inference about readers, and specifically about their linguistic proficiency has thus far been largely unutilized.

Here, we first introduce EyeScore, an independent measure of ESL proficiency which reflects the extent to which a learner's English reading patterns resemble those of native speakers. Second, we present a regression model which uses gaze features to predict the learner's scores on specific external proficiency tests. We address each of our tasks in two data regimes: \emph{Fixed Text}, which requires eyetracking training data for the specific sentences presented to the test taker, as well as the more general and challenging \emph{Any Text} regime, where the test taker is presented with arbitrary sentences for which no previous eyetracking data is available. To enable prediction mechanisms in both regimes, we utilize previously proposed gaze features, and develop new linguistically and psychologically motivated feature sets which capture the interaction between eye movements and linguistic properties of the text. 

We demonstrate the effectiveness of our approach via score comparison to standardized English proficiency tests. Our primary benchmark test, taken in lab by 145 ESL participants, are the grammar and listening sections of the Michigan English Test (MET) whose scores range from 0 to 50. EyeScore yields 0.5 Pearson's correlation to MET in the Fixed Text regime, and 0.48 in the Any Text regime. Our regression model for predicting MET scores from eye movement features obtains a correlation of 0.7 and a Mean Absolute Error (MAE) of 3.31 points in the Fixed Text regime, and 0.49 correlation and 4.11 MAE in the Any Text regime. Our results are substantially stronger compared to a baseline using only raw reading speed, and are reasonably close to correlations among traditional proficiency tests. These outcomes confirm the promise of the proposed methodology to reliably measure language proficiency.

This paper is structured as follows. Section \ref{sec:setup} describes the data and the experimental setup. In section \ref{sec:features} we delineate our feature sets for charactering eye movements in human reading. Section \ref{sec:eyescore} introduces EyeScore, a second language proficiency metric which is based on similarity of reading patterns to native speakers. In section \ref{sec:pred} we use eyetracking patterns to predict scores on MET and TOEFL. In section \ref{sec:relatedwork} we survey related work. Finally, we conclude and discuss future work in section \ref{sec:conclusion}.
 
\section{Experimental Setup}
\label{sec:setup}

Our study uses the dataset of eye movement records and English proficiency scores introduced in Berzak et al. \shortcite{berzak2017}\footnote{The data was collected under IRB approval, and all the participants provided written informed consent.	}, which we describe here in brief. The dataset contains gaze recordings of 37 native English speakers and 145 ESL speakers belonging to four native language backgrounds: 36 Chinese, 36 Japanese, 36 Portuguese and 37 Spanish. Participants were presented with free-form English sentences appearing as one-liners. To encourage attentive reading each sentence was followed by a yes/no comprehension question. During the experiment participants held a controller with buttons for indicating sentence reading completion and answering the sentence comprehension questions. Participants' eye movements were recorded using a desktop mount EyeLink 1000 eyetracker (SR Research) at a sampling rate of 1000Hz.

\subsection{Procedure and Reading Materials}
An experimental trial for a sentence starts with a presentation of a target circle at the center left of a blank screen. A 300ms fixation on this circle triggers a one-liner sentence on a new screen starting at the same location. After completing reading the sentence, participants are presented with the letter Q on a blank screen. A 300ms fixation on this letter triggers a question about the sentence on a new screen. Participants provide a yes/no answer to the question and are subsequently informed if they answered correctly. The first trial of the experiment was presented to familiarize participants with the experimental setup, and is discarded from the analysis.

Each participant read a total of 156 English sentences, randomly drawn from the Wall Street Journal Penn Treebank (WSJ-PTB) \cite{marcus1993}. The maximal sentence length was set to 100 characters, yielding an average sentence length of 11.4 words. All the sentences include the manual PTB annotations of POS tags \cite{santorini1990} and phrase structure trees, as well as Google universal POS tags \cite{petrov2012} and dependency trees obtained from the Universal Dependency Treebank (UDT) \cite{mcdonald2013}.

\subsection{Experimental Regimes}

Half of the 156 sentences presented to each participant belong to the \emph{Fixed Text} regime, and the other half belong to the \emph{Any Text} regime. Sentences from the two regimes were interleaved randomly and presented to all participants in the same order.

\textbf{Fixed Text} In this regime, all the participants read the same suite of 78 pre-selected sentences (900 words). The Fixed Text regime supports \emph{token-level} comparisons of reading patterns for specific words in the same contexts across readers. It enables the construction of a proficiency test which relies on a fixed battery of reading materials for which previous eyetracking data was collected.

\textbf{Any Text} In the second, Any Text regime, different participants read different sets of 78 sentences each (880 words on average). This regime generalizes the Fixed Text scenario; predicting reader characteristics in this regime requires formulating \emph{type-level} abstractions that would allow meaningful comparisons of reading patterns across different sentences. It corresponds to a proficiency test in which the sentences presented to the test taker are completely arbitrary, and no prior eyetracking data is available for them. 

\subsection{Standardized English Tests}

We use participants' performance on the Michigan English Test (MET) and TOEFL as external benchmarks of their English proficiency.

\textbf{Michigan English Test (MET)} Our primary indicator of English proficiency is the listening and grammar sections of the MET (Form-B), which were administered by Berzak et al. \shortcite{berzak2017} in-lab, and taken by all the 145 non-native participants upon completion of the reading experiment. The test has a total of 50 multiple choice questions, comprising 20 listening comprehension questions and 30 written grammar questions. The test score is computed as the number of correct answers for these questions, with possible scores ranging from 0 to 50. The mean MET score in the dataset is 41.46 (std 6.27).

\textbf{TOEFL} Berzak et al. \shortcite{berzak2017} also collected self-reported scores on the most recently taken official English proficiency test, which we use here as a secondary evaluation benchmark. We focus on the most commonly reported test, the TOEFL-iBT whose scores range from 0 to 120. We take into account only test results obtained less than four years prior to the experiment, yielding 33 participants. We sum the scores of the reading and listening sections of test, with a total possible score range of 0 to 60. In cases where participants reported only the overall score, we divided that score by two. We further augment this data with 20 participants who took the TOEIC Listening and Reading test within the same four years range, resulting in a total of 53 external proficiency scores. The TOEIC scores were converted to the TOEFL scale by fitting a third degree polynomial on an unofficial score conversion table\footnote{http://theedge.com.hk/conversion-table-for-toefl-ibt-pbt-cbt-tests/ Although both TOEFL and TOEIC are administered by the same company (ETS), to the best of our knowledge there is no publicly available official conversion table between the two tests.} between the tests. The converted scores were then divided by two. Henceforth we refer to both TOEFL-iBT and TOEIC scores converted to TOEFL-iBT scale as TOEFL scores. The mean TOEFL score is 47.6 (std 9.55). The Pearson's $r$ correlation between the TOEFL and MET scores in the dataset is 0.74. 

\subsection{Data Split}

\begin{figure}
  \centering
    \includegraphics[width=0.45\textwidth]{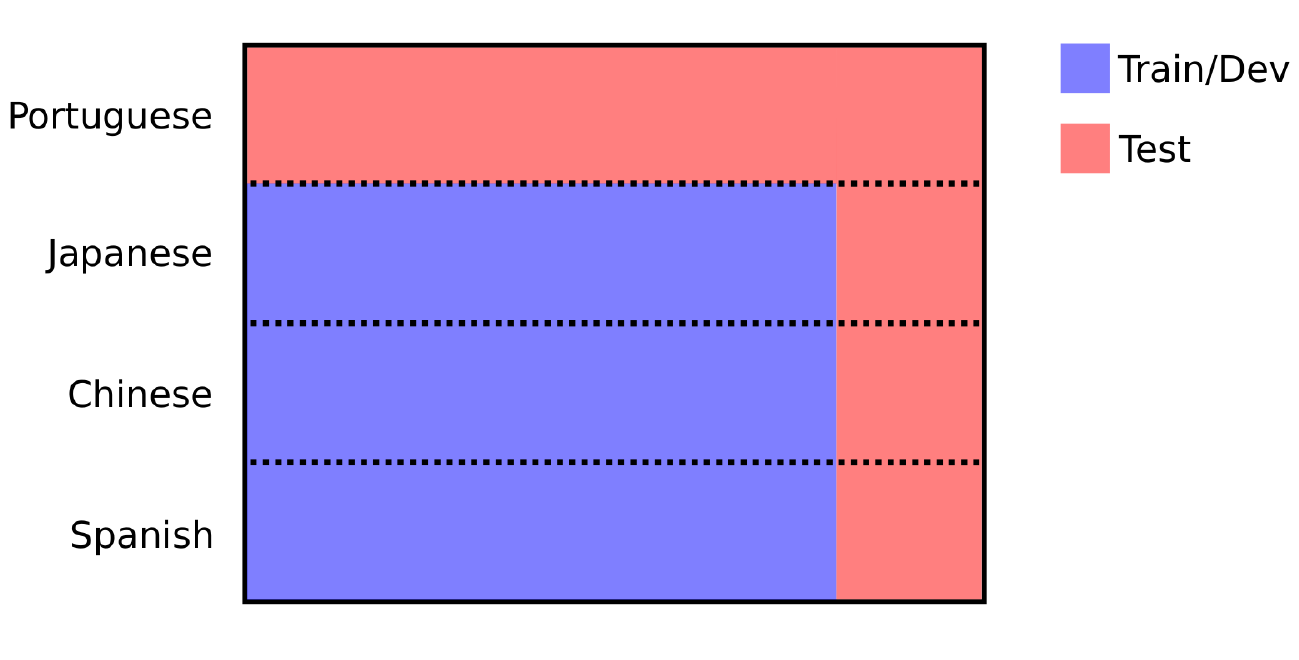}
  \caption{Illustration of the data split for MET into a training/development set (88 participants) and a test set (57 participants).}\label{datasplit-fig}
\end{figure}

We divide the ESL speakers into training/development and test sets in the following manner. For MET, we split our 145 ESL participants into a training/development set of 88 participants and a test set of 57 participants. The test set consists of an entire held out native language -- 36 speakers of Portuguese -- as well as 7 participants randomly sampled from each of the remaining three native languages. Our test set is thus particularly challenging due to the large fraction of participants belonging to the held out language, a design which emphasizes generalization to language learner populations which are not part of the training set. Figure \ref{datasplit-fig} presents a schematic overview of our MET split. For TOEFL, due to the limited available data, in Section \ref{sec:eyescore} we report EyeScore correlations for all the 53 test takers, and in Section \ref{sec:pred} we perform regression experiments using leave-one-out cross validation. 

\section{Eye Movement Features}
\label{sec:features}

In order to capture behavioral psycholinguistic traces of language proficiency we utilize several linguistically and psychologically motivated feature representations of eye movements in reading. We include features introduced in prior work (see Words in Fixed Context and Syntactic Clusters \cite{berzak2017}) as well as newly developed feature sets (see Word Property Coefficients and Transitions). All our features rely on the well established division of gaze trajectories into fixations (stops) and saccades (movements between fixations) that characterizes human reading \cite{rayner1998}. 

Our fixation based features make use of several standard metrics of fixation times, defined below.
\begin{itemize}
\item \emph{First Fixation duration (FF)} Duration of the first fixation on a word.
\item \emph{First Pass duration (FP)} Time spent from first entering a word to first 
leaving it (including re-fixations within the word). 
\item \emph{Total Fixation duration (TF)} The sum of all fixation times on a word.
\item \emph{Regression Path duration (RP)} Time from first entering a word until proceeding to its right.
\end{itemize}

Our feature sets are divided into two groups. The first group consists of type-level features, applicable both in the Any Text and Fixed Text regimes. The second group of feature sets is token-based and can be extracted only in the Fixed Text regime, because it presupposes the same textual input for all participants.
 
\subsection{Type-Level Features}

\subsubsection*{Word Property Coefficients (WP-Coefficients)} 

This new feature set quantifies the influence of three key word characteristics on reading times of individual readers: word length, word frequency and surprisal. The last measures the difficulty of processing a word in a sentence \cite{hale2001,levy2008}, and is defined as its negative log probability given a sentential context:
\begin{equation}
surprisal(w_{i}|w_{1...i-1}) = -\log(w_{i}|w_{1...i-1})
\end{equation}
In the reading literature, these three characteristics were suggested as the most prominent linguistic factors influencing word reading times \citep[e.g.][]{inhoff1986,rayner1996,pollatsek2008,kliegl2004,rayner2004,rayner2011,smith2013,luke2016limits}; whereby longer, less frequent and contextually less predictable words are fixated longer.

To derive this feature set, we measure length as the number of characters in the word. Word (log) frequencies are obtained from the BLLIP-WSJ corpus \cite{bllip2000}. Estimates of surprisal are obtained from a trigram language model with Chen and Goodman's modified Kneser-Ney smoothing trained on the BLLIP-WSJ using SRILM \cite{stolcke2002}. We then fit for each participant four regression models that use these three word characteristics to predict the word's raw FF, FP, TF and RP durations. The regression models are fitted using Ordinary Least Squares (OLS). We also train a logistic regression model for predicting word skips. Finally, we extract the weights and intercepts of these models and encode them as features. As each of the five models has three coefficients and one intercept term, the resulting WP-Coefficients feature set has 20 features.

\subsubsection*{Syntactic Clusters (S-Clusters)}

Following Berzak et al. \shortcite{berzak2017}, we extract average word reading times clustered by POS tags and syntactic functions. We utilize three metrics of reading times, FF, FP and TF durations. We then cluster words according to three types of syntactic criteria, Google Universal POS tags, PTB POS tags, and the syntactic function label of the word to its head word. To derive the feature set, we average the word fixation times of each cluster. An example of an S-Cluster feature is the average TF duration for words with the PTB POS tag DT. We take into account only cluster labels that appear at least once in the reading input of all the participants, yielding a total of 312 S-Clusters features in the Fixed Text regime. In the Any Text regime we obtain 156 S-Clusters features for MET and 165 S-Clusters features for TOEFL.

\subsection{Token-Level Features}

\subsubsection*{Transitions}

Transitions is a new feature set which summarizes the sequence of saccades between words in a sentence. Given a sentence with $n$ words, we construct an $n \times n$ matrix $T$. A matrix entry $t_{i,j}$ records the number of saccades whose launch site falls within word $i$ and landing site falls within word $j$. With a total of 11,616 possible transitions in the Fixed Text sentences, the resulting feature set contains 9,077 features with a non-zero value for at least one participant for MET, and 8,132 such features for TOEFL.

\subsubsection*{Words in Fixed Context (WFC)}

This feature set was previously used in Berzak et al. \shortcite{berzak2017} and consists of reading times for words within fixed contexts. We extract FP and TF durations for the 900 words in the Fixed Text sentences, resulting in a total of 1,800 WFC features.

\section{English Proficiency Scoring Based on Eye Movements in Reading}
\label{sec:eyescore}

We hypothesize that language proficiency influences the way that learners process a second language, which in turn will be reflected in eye movement patterns in reading. Specifically, we propose to examine whether the more proficient is an ESL learner, the more similar are their reading patterns to those of native English speakers. We operationalize the notion of native-like reading in the following manner. First, given a feature representation of choice and a dataset $D$ comprising ESL learners $D_{L2}$ and native speakers $D_{L1}$ we Z score each feature in $D$ using a Z scaler derived from $D_{L2}$. We then obtain a prototype feature vector of native reading $v_{L1}$ by averaging the feature vectors of the native speakers. 
\begin{equation}
v_{L1} = \frac{1}{|D_{L1}|}\sum_{y \in D_{L1}}{v_y}
\end{equation}
Finally, we obtain an eyetracking based proficiency score of an ESL learner by computing the cosine similarity of their feature vector to the native reading prototype. Hereafter we refer to this measure as EyeScore.
\begin{equation}
EyeScore_{y \in D_{L2}} = \frac{v_{y} \cdot v_{L_1}}{\Vert v_{y}\Vert \Vert v_{L1}\Vert }
\end{equation}

\textbf{Reading Speed Normalization} To reduce bias towards fast readers, the feature representations used for Eyescore are normalized to be invariant to the reading speed of the participant. Specifically, for the S-Clusters and WFC feature sets we follow the normalization procedure of Berzak et al. \shortcite{berzak2017}, where for a given participant, the reading time of a word $w_{i}$ according to a fixation metric $M$ is normalized by $S_{M,C}$, the metric's fixation time per word in the linguistic context $C$:
\begin{equation}
S_{M,C} = \frac{1}{|C|}\sum_{w \in C}{M_{w}}
\end{equation}
The linguistic context is defined as the surrounding sentence in the Fixed Text regime, and the entire textual input in the Any Text regime. The normalized fixation time is then obtained as:
\begin{equation}
Mnorm_{w_{i}} = \frac{M_{w_{i}}}{S_{M,C}}
\end{equation}
For the WC-Coefficients features we take into account only the 15 model coefficients, and omit the 5 intercept features which capture the reading speed of the participant. Finally, we also normalize the Transitions features matrix $T$ by the total number of saccades in the sentence to obtain 
$T_{norm}$ in which $\sum_{i,j}t_{norm_{i,j}} = 1$.

\subsection{Correlation with MET and TOEFL}

We evaluate the ability of EyeScore to capture language proficiency by comparing it against our two external proficiency tests, MET and TOEFL. Table \ref{table:eyescore} presents the Pearson's $r$ correlation of EyeScore with MET and TOEFL for the feature sets described in section \ref{sec:features} using the MET training/development set and all the participants who took TOEFL. 

\begin{table}[ht!]
\resizebox{\columnwidth}{!}{
\begin{tabular}{|l|ll|ll|}
 \multicolumn{1}{c}{} & \multicolumn{2}{c}{\bf{MET}} & \multicolumn{2}{c}{\bf{TOEFL}} \\ \hline
\bf Features 	& Fixed	& Any	& Fixed	& Any \\ \hline
Reading Speed 	& 0.28	& 0.27 	& 0.15 	& 0.13	\\ \hline
WP-Coefficients	& 0.38 	& 0.37	& 0.21	& 0.13 \\
S-Clusters	& 0.45 	&\bf0.48& 0.50	&\bf0.45 \\  \hline
Transitions 	& 0.45 	& NA 	& 0.44  & NA \\ 
WFC 		&\bf0.50 & NA 	&\bf0.54& NA\\ \hline
\end{tabular}
}
\caption{Pearson's $r$ of EyeScore for different feature sets with MET (training/development set, 88 participants) and TOEFL (all 53 participants). Fixed denotes the Fixed Text regime in which all the participants read the same sentences, and Any denotes the Any Text regime where different readers read different sentences. 
}\label{table:eyescore}
\end{table} 

\begin{figure}[t!]
  \centering
    \includegraphics[width=0.48\textwidth]{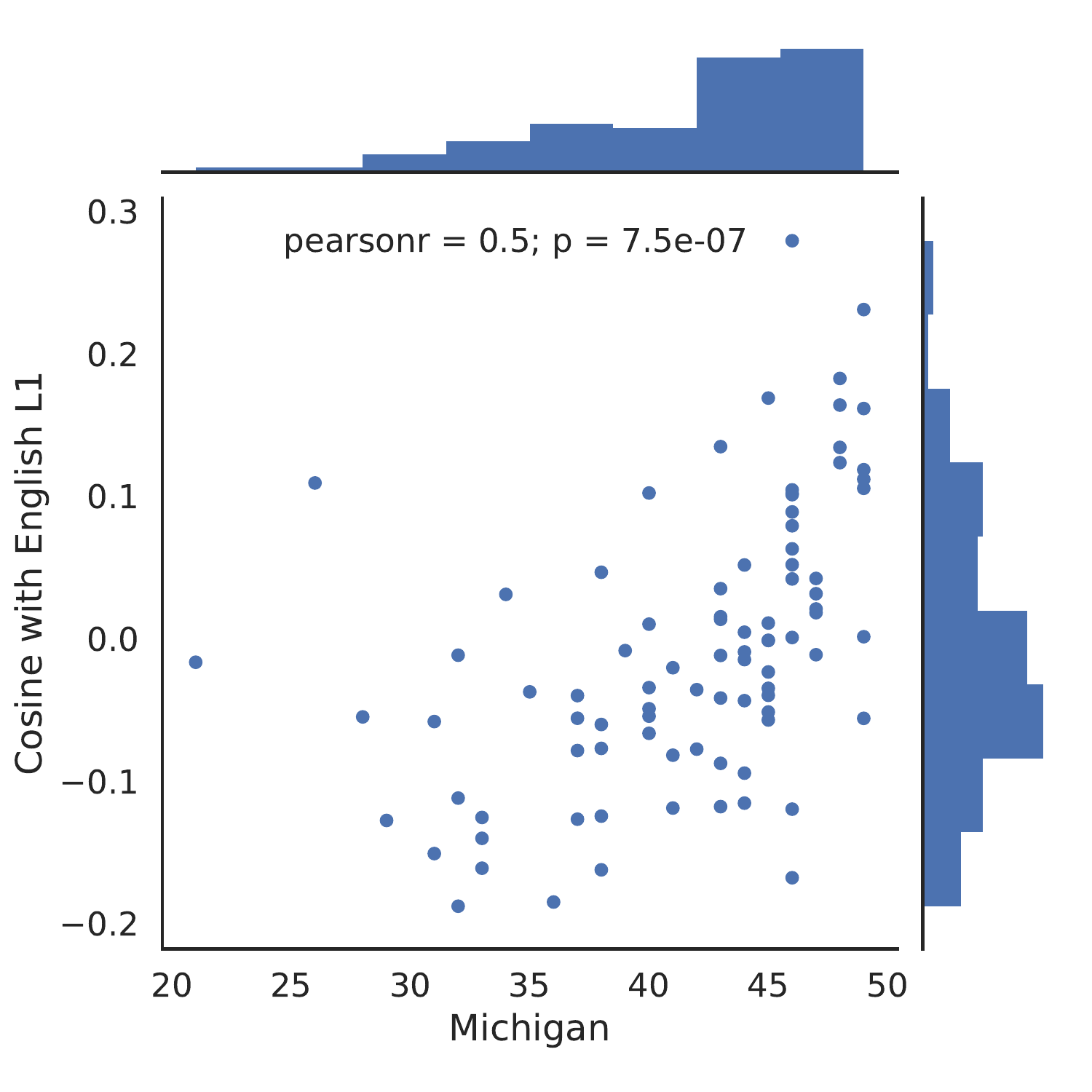}
  \caption{Comparison of MET (training/development set, 88 participants) with EyeScore using Words in Fixed Context (WFC) features in the Fixed Text regime.}\label{corr-plot}
\end{figure}

The strongest correlations, 0.5 for MET and 0.54 for TOEFL, are obtained in the Fixed Text regime using the WFC features. This outcome confirms the effectiveness reading time comparisons when the presented sentences are shared across participants. To illustrate the quality of this result, Figure \ref{corr-plot} presents a comparison of EyeScore and MET scores in the Fixed Text and WFC features setup. We further note good performance of the Transitions and S-Clusters features in this regime across both proficiency tests. The strongest performance in the Any Text regime is obtained using the S-Clusters features, yielding 0.48 correlation with MET and 0.45 correlation with TOEFL. These results are competitive with the WFC feature set in the Fixed Text regime, suggesting that reliable EyeScores can be obtained even when no prior eyetracking data is available for the sentences presented to the test taker. 

In order to contextualize the correlations obtained with the EyeScore approach, we first compare our results to raw reading speed, an informative baseline which does not rely on eyetracking. EyeScore substantially outperforms this baseline for nearly all the feature sets on both MET and TOEFL, clearly showing the benefit of eye movement information for our task. Next, we consider possible upper bounds for our correlations. While obtaining such upper bounds is challenging, we can use correlations between different traditional standardized proficiency tests as informative reference points. First, as mentioned previously, in our dataset the MET and reported TOEFL scores have a Pearson's $r$ correlation of 0.74. We further note an external study conducted by the testing company Education First (EF) which measured the correlation of their flagship standardized English proficiency test EFSET-PLUS with TOEFL-iBT \cite{efreport2015}. Using 384 participants who took both tests, the study found a Pearson's $r$ of 0.63 for the reading comprehension and 0.69 for the listening comprehension sections of these tests. Despite the radical difference of our testing methodology, our strongest feature sets obtain rather competitive results relative to these correlations, further strengthening the evidence for the ability of our approach to capture language proficiency.

\section{Predicting Performance on MET and TOEFL}
\label{sec:pred}

\begin{table*}[ht!]
  \centering
%\resizebox{\textwidth}{!}{
\begin{tabular}{|l|llll|llll|}
\multicolumn{1}{c}{} 		& \multicolumn{4}{c}{ \bf MET }	&\multicolumn{4}{c}{ \bf TOEFL }\\ \cline{2-9}
\multicolumn{1}{c}{} & \multicolumn{2}{|c}{Fixed} & \multicolumn{2}{c}{Any} & \multicolumn{2}{|c}{Fixed} & \multicolumn{2}{c|}{Any} \\ \cline{2-9}
 \multicolumn{1}{l}{\bf Features} &  \multicolumn{1}{|c}{$r$}& \multicolumn{1}{c}{MAE} 	& \multicolumn{1}{c}{$r$}	&  \multicolumn{1}{c}{MAE}	& \multicolumn{1}{|c}{$r$}	& \multicolumn{1}{c}{MAE} & \multicolumn{1}{c}{$r$} & \multicolumn{1}{c|}{MAE} \\ \hline
Reading Speed 	& 0.27 	& 4.58 	& 0.24 	& 4.62 		& 0.09	& 7.92 	& 0.06	& 7.96 \\ \hline
WP-Coefficients	& 0.43	& 4.11	& 0.44	& 4.14		& 0.34 	& 7.76	& 0.31	&\bf 7.49 \\
S-Clusters 	& 0.56 	& 3.87	&\bf0.49& \bf4.11 	& \bf0.55	& 7.45	&\bf0.50& 7.76 \\ \hline
Transitions 	& 0.52	& 3.93	& NA 	& NA 		& 0.38	& 7.11	& NA 	& NA \\
WFC 		&\bf0.70&\bf3.31& NA 	& NA 		&0.50&\bf6.68& NA 	& NA \\ \hline
\end{tabular}
\caption{Pearson's $r$ and Mean Absolute Error (MAE) for prediction of MET scores (test set, 57 participants) and TOEFL scores (leave-one-out cross validation, all 53 participants) from eye movement patterns in reading. We consider two baselines which do not use eyetracking information: (1) the average proficiency score in the training set, which yields 4.82 MAE on MET and 8.29 MAE on TOEFL, and (2) the reading speed of the participant. 
}\label{table:testpred}
%}
\end{table*}

In section \ref{sec:eyescore} we introduced EyeScore as an independent metric of language proficiency which is based on eye movements during reading. Here, we examine whether eye movements can also be used to explicitly \emph{predict} the performance of participants on specific external standardized language proficiency tests. This task is of practical value for development of predictive tools for standardized proficiency tests, and constitutes an alternative framework for studying the relevance of eye movement patterns in reading to language proficiency.   

To address this task, we use Ridge regression to predict overall scores on an external proficiency test from eye movement features in reading. The model parameters $\theta$ are obtained by minimizing the following loss objective:
\begin{equation}
\sum_{i}{(y_{i} - \theta\cdot f(x_{i}))^2} + \lambda\lVert\theta\rVert_2^2
\end{equation}
where $y_{i}$ is a participant's test score, $x_{i}$ is their eye movement record, and $f(x_{i})$ are the extracted eye movement features. To calibrate the model with respect to native English speakers, we augment each training set $D_{L2_{tr}}$ with the group of 37 native speakers $D_{L1}$ whose proficiency scores are assigned to the maximum grade of the respective test (50 for MET and 60 for TOEFL)\footnote{Our experiments on the training/development set indicate that this training data augmentation step leads in most cases to improved regression performance.}. Based on MET performance on the train/dev set, the features used for predicting scores on both tests are not normalized for speed\footnote{We note that in line with the low correlation of reading speed with TOEFL, speed normalized features tend to be better predictors of TOEFL scores, obtaining $r$ 0.59 and MAE 6.47 with WFC features in the Fixed Text regime, and $r$ 0.58 and MAE 7.19 with S-Clusters in the Any Text regime.}. As a preprocessing step, we fit a Z scaler for each feature using the ESL participants in the training set, and apply it to all the participants in the training and test sets.

\subsection*{Results}

We evaluate prediction accuracy using Pearson's $r$ and Mean Absolute Error (MAE) from the true proficiency test scores. The $\lambda$ parameter for MET is optimized for MAE on 10 fold cross validation within the training/development set. For TOEFL, which has a relatively small number of participants, we report results on leave-one-out cross validation with $\lambda$ set to 1. 

\begin{figure}[ht!]
  \centering
    \includegraphics[width=0.48\textwidth]{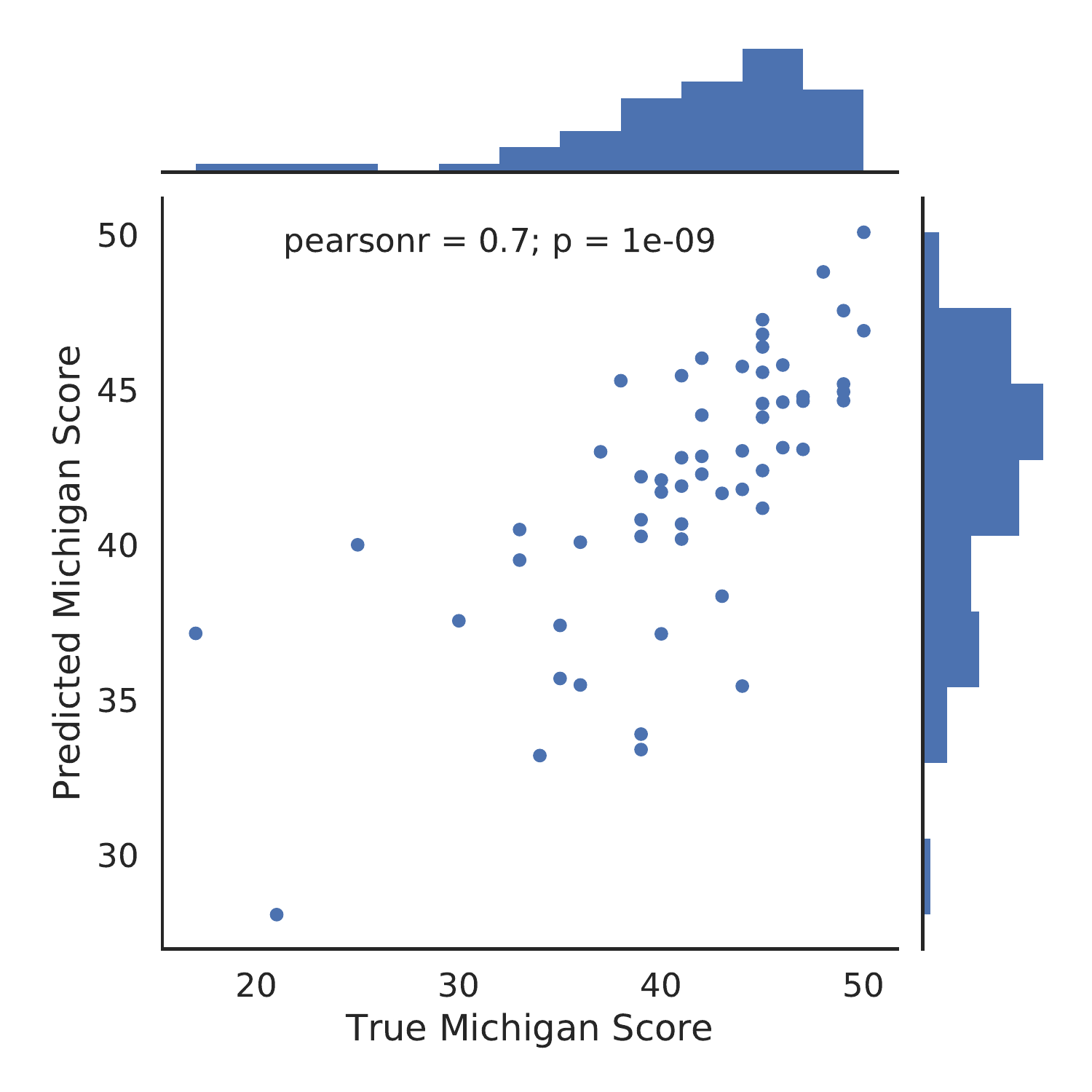}
  \caption{Comparison of MET scores (test set, 57 participants) with predicted MET scores using Words in Fixed Context (WFC) eye movement features in the Fixed Text regime.}\label{pred-plot}
\end{figure}

Table \ref{table:testpred} presents the results for both proficiency tests. We consider two baselines; the first is assigning all test set participants with the average score of the training participants. This baseline yields an MAE of 4.82 on MET and 8.29 on TOEFL. The second baseline uses reading speed as the sole feature for prediction. In all cases, our eyetracking based features outperform the average score and reading speed baselines. 

The performance of the different feature sets is in most cases consistent across the two proficiency tests and is largely in line with the correlations of EyeScore reported in Table \ref{table:eyescore}.
Similarly to the EyeScore outcomes, the best performance in the Fixed Text regime is obtained using the WFC feature set, with a Pearson's $r$ of 0.7 and MAE of 3.31 for MET. This result is highly competitive with correlations between different standardized English proficiency tests. Figure \ref{pred-plot} depicts a comparison between MET scores and our MET predictions in this setup. On TOEFL, WFC features obtain the strongest MAE of 6.68, while S-Clusters have a higher $r$ coefficient of 0.55. 

In the Any Text regime, differently from EyeScore, we obtain comparable results for the S-Clusters and WP-Coefficients feature sets. Overall, the improvements of both feature sets over the baselines in the Any Text regime further support the ability of type-level features to generalize the task of language proficiency prediction to arbitrary sentences.

\section{Related Work}
\label{sec:relatedwork}

Our work lies on the intersection of language proficiency assessment, second language acquisition (SLA), the psychology of reading and NLP. Automated language proficiency assessment from free-form linguistic performance has been studied mainly in language \emph{production} \cite{dikli2006,williamson2009,shermis2013}. Over the past several decades, multiple essay and speech scoring systems have been developed for learner language using a wide range of linguistically motivated feature sets \citep[e.g.][]{lonsdale2003,landauer2003,xi2008,yannakoudakis2011}. Some of these systems have been deployed in official language proficiency tests, for example the \emph{e-rater} essay scoring system \cite{attali2004} used in TOEFL \cite{ramineni2012}. While this line of work focuses on assessment of language production, here we introduce and address for the first time automated language assessment during online language \emph{comprehension}.

In SLA, there has been considerable interest in eyetracking, where studies have mostly focused on controlled experiments examining processing of specific linguistic phenomena such as syntactic ambiguities, cognates and idioms \cite{dussias2010,roberts2013}. A notable exception is \cite{copl15} who used free-form reading to study differences in fixation times and saccade lengths between native and non-native readers. Our work also adopts broad coverage analysis of reading patterns, which we use to formulate predictive models of language proficiency.

Our study draws on a large body of work in the psychology of reading \citep[see][for overview]{rayner1998,rayner2012} which has suggested that eye movement patterns during reading are systematically influenced by a broad range of linguistic characteristics of the text, and reflect how readers mentally engage with the text \citep[][among many others]{frazier1982,rayner1989,reichle1998,engbert2005,demberg2008,reichle2009,levy2009}. Prior work on reading has also demonstrated that gaze provides valuable information about various characteristics of the reader and their cognitive state. For example, Reichle et at. \shortcite{reichle2010} have shown that eye movement patterns are categorically different in attentive versus mindless reading. In Rello and Ballesteros \shortcite{rello2015} eye movements were used to distinguish between readers with and without dyslexia. Berzak et al. \shortcite{berzak2017} collected the dataset used in our work and used it to predict the first language of non-native English readers from gaze. We build on these studies to motivate our task and design feature representations which encode linguistic factors known to affect the human reading process.

Related work in NLP developed predictive models of reading times in reading of free-form text \citep[e.g.][]{nilsson2009,hara2012,hahn2016}. In a complementary vein, eyetracking signal has been used for linguistic annotation tasks such as POS tagging \cite{barrett2015pos,barrett2016} and prediction of syntactic functions \cite{barrett2015functions}. Both lines of investigation provide further evidence for the tight interaction between eye movements and linguistic properties of the text, which we leverage in our work for inference about the linguistic knowledge of the reader.

\section{Conclusion and Discussion}
\label{sec:conclusion}

We present a novel approach for automated assessment of language proficiency which relies on eye movements during reading of free-form text. Our EyeScore test captures the similarity of language learners' gaze patterns to those of native speakers, and correlates well with the standardized tests MET and TOEFL. A second variant of our approach accurately predicts participants' scores on these two tests. To the best of our knowledge, the proposed framework is the first proof-of-concept for a system which utilizes eyetracking to measure linguistic ability.

In future work, we plan to extend the analysis of the validity and consistency of our approach, and further explore its applications for language proficiency evaluation. In particular, we will examine the impact of factors that can undermine the validity of language proficiency tests, such as test specific training, familiarity with the evaluation system's features \cite{powers2002}, and cheating via unauthorized prior access to test materials. Since participants are less likely to be able to manipulate their eye movements in an informed and systematic manner---readers are generally not even aware that their eye movements are saccadic---and since our test can be performed on arbitrary sentences, we expect it to be robust to prior exposure to the test materials and testing methodology.
We will further study the consistency of our scores for repeated tests by the same participants. A preliminary split-half analysis indicates that eyetracking based scores are expected to be highly consistent across tests. Finally, our approach can be combined with traditional proficiency testing methodologies, whereby gaze will be recorded while the participant is taking a standardized language proficiency test. This will enable developing novel approaches to language proficiency assessment which will integrate task based performance with real time monitoring of cognitive and linguistic processing.

\section*{Acknowledgments}
This material is based upon work supported in part by the Center for Brains, Minds, 
and Machines (CBMM), funded by NSF STC award CCF-1231216.

%\bibliography{naaclhlt2018}
\bibliography{naaclhlt2018}

\begin{thebibliography}{}
\expandafter\ifx\csname natexlab\endcsname\relax\def\natexlab#1{#1}\fi

\bibitem[{Attali and Burstein(2004)}]{attali2004}
Yigal Attali and Jill Burstein. 2004.
\newblock Automated essay scoring with e-rater{\textregistered} v. 2.0.
\newblock {\em ETS Research Report Series\/} 2004(2).

\bibitem[{Barrett et~al.(2016)Barrett, Bingel, Keller, and
  S{\o}gaard}]{barrett2016}
Maria Barrett, Joachim Bingel, Frank Keller, and Anders S{\o}gaard. 2016.
\newblock Weakly supervised part-of-speech tagging using eye-tracking data.
\newblock In {\em ACL\/}. volume~2, pages 579--584.

\bibitem[{Barrett and S{\o}gaard(2015{\natexlab{a}})}]{barrett2015pos}
Maria Barrett and Anders S{\o}gaard. 2015{\natexlab{a}}.
\newblock Reading behavior predicts syntactic categories.
\newblock In {\em CoNLL\/}. pages 345--349.

\bibitem[{Barrett and S{\o}gaard(2015{\natexlab{b}})}]{barrett2015functions}
Maria Barrett and Anders S{\o}gaard. 2015{\natexlab{b}}.
\newblock Using reading behavior to predict grammatical functions.
\newblock In {\em Proceedings of the Sixth Workshop on Cognitive Aspects of
  Computational Language Learning\/}. pages 1--5.

\bibitem[{Berzak et~al.(2017)Berzak, Nakamura, Flynn, and Katz}]{berzak2017}
Yevgeni Berzak, Chie Nakamura, Suzanne Flynn, and Boris Katz. 2017.
\newblock Predicting native language from gaze.
\newblock In {\em ACL\/}. pages 541--551.

\bibitem[{Charniak et~al.(2000)Charniak, Blaheta, Ge, Hall, Hale, and
  Johnson}]{bllip2000}
Eugene Charniak, Don Blaheta, Niyu Ge, Keith Hall, John Hale, and Mark Johnson.
  2000.
\newblock {BLLIP} 1987-89 {WSJ} corpus release 1.
\newblock {\em Linguistic Data Consortium, Philadelphia\/} 36.

\bibitem[{Cop et~al.(2015)Cop, Drieghe, and Duyck}]{copl15}
Uschi Cop, Denis Drieghe, and Wouter Duyck. 2015.
\newblock Eye movement patterns in natural reading: A comparison of monolingual
  and bilingual reading of a novel.
\newblock {\em PLOS ONE\/} 10(8):1--38.

\bibitem[{Demberg and Keller(2008)}]{demberg2008}
Vera Demberg and Frank Keller. 2008.
\newblock Data from eye-tracking corpora as evidence for theories of syntactic
  processing complexity.
\newblock {\em Cognition\/} 109(2):193--210.

\bibitem[{Dikli(2006)}]{dikli2006}
Semire Dikli. 2006.
\newblock An overview of automated scoring of essays.
\newblock {\em The Journal of Technology, Learning and Assessment\/} 5(1).

\bibitem[{Dussias(2010)}]{dussias2010}
Paola~E Dussias. 2010.
\newblock Uses of eye-tracking data in second language sentence processing
  research.
\newblock {\em Annual Review of Applied Linguistics\/} 30:149--166.

\bibitem[{Engbert et~al.(2005)Engbert, Nuthmann, Richter, and
  Kliegl}]{engbert2005}
Ralf Engbert, Antje Nuthmann, Eike~M Richter, and Reinhold Kliegl. 2005.
\newblock Swift: a dynamical model of saccade generation during reading.
\newblock {\em Psychological Review\/} 112(4):777.

\bibitem[{Frazier and Rayner(1982)}]{frazier1982}
Lyn Frazier and Keith Rayner. 1982.
\newblock Making and correcting errors during sentence comprehension: Eye
  movements in the analysis of structurally ambiguous sentences.
\newblock {\em Cognitive Psychology\/} 14(2):178--210.

\bibitem[{Hahn and Keller(2016)}]{hahn2016}
Michael Hahn and Frank Keller. 2016.
\newblock Modeling human reading with neural attention.
\newblock In {\em EMNLP\/}. pages 85--95.

\bibitem[{Hale(2001)}]{hale2001}
John Hale. 2001.
\newblock A probabilistic {E}arley parser as a psycholinguistic model.
\newblock In {\em NAACL\/}. Association for Computational Linguistics, pages
  1--8.

\bibitem[{Hara et~al.(2012)Hara, Mochihashi, Kano, and Aizawa}]{hara2012}
Tadayoshi Hara, Daichi Mochihashi, Yoshinobu Kano, and Akiko Aizawa. 2012.
\newblock Predicting word fixations in text with a {CRF} model for capturing
  general reading strategies among readers.
\newblock In {\em Proceedings of the 1st Workshop on Eye-tracking and Natural
  Language Processing\/}. pages 55--70.

\bibitem[{IELTS(2017)}]{wiki:ielts}
IELTS. 2017.
\newblock
  \href{https://en.wikipedia.org/wiki/International_English_Language_Testing_System}{International
  {E}nglish language testing system --- {W}ikipedia{,} the free encyclopedia}.
\newblock Online; accessed November 2017.
\newblock
  \url{https://en.wikipedia.org/wiki/International_English_Language_Testing_System}.

\bibitem[{Inhoff and Rayner(1986)}]{inhoff1986}
Albrecht~Werner Inhoff and Keith Rayner. 1986.
\newblock Parafoveal word processing during eye fixations in reading: Effects
  of word frequency.
\newblock {\em Perception \& Psychophysics\/} 40(6):431--439.

\bibitem[{Kliegl et~al.(2004)Kliegl, Grabner, Rolfs, and Engbert}]{kliegl2004}
Reinhold Kliegl, Ellen Grabner, Martin Rolfs, and Ralf Engbert. 2004.
\newblock Length, frequency, and predictability effects of words on eye
  movements in reading.
\newblock {\em European Journal of Cognitive Psychology\/} 16(1-2):262--284.

\bibitem[{Landauer(2003)}]{landauer2003}
Thomas~K Landauer. 2003.
\newblock Automated scoring and annotation of essays with the {I}ntelligent
  {E}ssay {A}ssessor.
\newblock {\em Automated essay scoring: A crossdisciplinary perspective\/} .

\bibitem[{Levy(2008)}]{levy2008}
Roger Levy. 2008.
\newblock Expectation-based syntactic comprehension.
\newblock {\em Cognition\/} 106(3):1126--1177.

\bibitem[{Levy et~al.(2009)Levy, Bicknell, Slattery, and Rayner}]{levy2009}
Roger Levy, Klinton Bicknell, Tim Slattery, and Keith Rayner. 2009.
\newblock Eye movement evidence that readers maintain and act on uncertainty
  about past linguistic input.
\newblock {\em Proceedings of the National Academy of Sciences\/}
  106(50):21086--21090.

\bibitem[{Lonsdale and Strong-Krause(2003)}]{lonsdale2003}
Deryle Lonsdale and Diane Strong-Krause. 2003.
\newblock Automated rating of {ESL} essays.
\newblock In {\em Proceedings of the HLT-NAACL Workshop on Building Educational
  Applications using Natural Language Processing - Volume 2\/}. Association for
  Computational Linguistics, pages 61--67.

\bibitem[{Luecht(2015)}]{efreport2015}
Richard~M Luecht. 2015.
\newblock
  \href{https://www.efset.org/research/~/media/centralefcom/efset/pdf/EFSET_TOEFL_correlational_report_Sep_v1.pdf}{{EFSET
  PLUS - TOEFL} i{BT} correlation study report}.
\newblock
  \url{https://www.efset.org/research/~/media/centralefcom/efset/pdf/EFSET_TOEFL_correlational_report_Sep_v1.pdf}.

\bibitem[{Luke and Christianson(2016)}]{luke2016limits}
Steven~G Luke and Kiel Christianson. 2016.
\newblock Limits on lexical prediction during reading.
\newblock {\em Cognitive Psychology\/} 88:22--60.

\bibitem[{Marcus et~al.(1993)Marcus, Marcinkiewicz, and Santorini}]{marcus1993}
Mitchell~P Marcus, Mary~Ann Marcinkiewicz, and Beatrice Santorini. 1993.
\newblock Building a large annotated corpus of {E}nglish: The {P}enn
  {T}reebank.
\newblock {\em Computational Linguistics\/} 19(2):313--330.

\bibitem[{McDonald et~al.(2013)McDonald, Nivre, Quirmbach-Brundage, Goldberg,
  Das, Ganchev, Hall, Petrov, Zhang, T{\"a}ckstr{\"o}m et~al.}]{mcdonald2013}
Ryan McDonald, Joakim Nivre, Yvonne Quirmbach-Brundage, Yoav Goldberg, Dipanjan
  Das, Kuzman Ganchev, Keith Hall, Slav Petrov, Hao Zhang, Oscar
  T{\"a}ckstr{\"o}m, et~al. 2013.
\newblock Universal dependency annotation for multilingual parsing.
\newblock In {\em ACL\/}.

\bibitem[{Nilsson and Nivre(2009)}]{nilsson2009}
Mattias Nilsson and Joakim Nivre. 2009.
\newblock Learning where to look: Modeling eye movements in reading.
\newblock In {\em Proceedings of the Thirteenth Conference on Computational
  Natural Language Learning\/}. Association for Computational Linguistics,
  pages 93--101.

\bibitem[{Petrov et~al.(2012)Petrov, Das, and McDonald}]{petrov2012}
Slav Petrov, Dipanjan Das, and Ryan McDonald. 2012.
\newblock A universal part-of-speech tagset.
\newblock In {\em LREC\/}.

\bibitem[{Pollatsek et~al.(2008)Pollatsek, Juhasz, Reichle, Machacek, and
  Rayner}]{pollatsek2008}
Alexander Pollatsek, Barbara~J Juhasz, Erik~D Reichle, Debra Machacek, and
  Keith Rayner. 2008.
\newblock Immediate and delayed effects of word frequency and word length on
  eye movements in reading: a reversed delayed effect of word length.
\newblock {\em Journal of Experimental Psychology: Human Perception and
  Performance\/} 34(3):726.

\bibitem[{Powers et~al.(2002)Powers, Burstein, Chodorow, Fowles, and
  Kukich}]{powers2002}
Donald~E Powers, Jill~C Burstein, Martin Chodorow, Mary~E Fowles, and Karen
  Kukich. 2002.
\newblock Stumping e-rater: challenging the validity of automated essay
  scoring.
\newblock {\em Computers in Human Behavior\/} 18(2):103--134.

\bibitem[{Ramineni et~al.(2012)Ramineni, Trapani, Williamson, Davey, and
  Bridgeman}]{ramineni2012}
Chaitanya Ramineni, Catherine~S Trapani, David~M Williamson, Tim Davey, and
  Brent Bridgeman. 2012.
\newblock Evaluation of the e-rater{\textregistered} scoring engine for the
  {TOEFL}{\textregistered} independent and integrated prompts.
\newblock {\em ETS Research Report Series\/} 2012(1).

\bibitem[{Rayner(1998)}]{rayner1998}
Keith Rayner. 1998.
\newblock Eye movements in reading and information processing: 20 years of
  research.
\newblock {\em Psychological Bulletin\/} 124(3):372.

\bibitem[{Rayner et~al.(2004)Rayner, Ashby, Pollatsek, and
  Reichle}]{rayner2004}
Keith Rayner, Jane Ashby, Alexander Pollatsek, and Erik~D Reichle. 2004.
\newblock The effects of frequency and predictability on eye fixations in
  reading: implications for the ez reader model.
\newblock {\em Journal of Experimental Psychology: Human Perception and
  Performance\/} 30(4):720.

\bibitem[{Rayner and Frazier(1989)}]{rayner1989}
Keith Rayner and Lyn Frazier. 1989.
\newblock Selection mechanisms in reading lexically ambiguous words.
\newblock {\em Journal of Experimental Psychology: Learning, Memory, and
  Cognition\/} 15(5):779.

\bibitem[{Rayner et~al.(2012)Rayner, Pollatsek, Ashby, and
  Clifton~Jr}]{rayner2012}
Keith Rayner, Alexander Pollatsek, Jane Ashby, and Charles Clifton~Jr. 2012.
\newblock {\em Psychology of reading\/}.
\newblock Psychology Press.

\bibitem[{Rayner et~al.(2011)Rayner, Slattery, Drieghe, and
  Liversedge}]{rayner2011}
Keith Rayner, Timothy~J Slattery, Denis Drieghe, and Simon~P Liversedge. 2011.
\newblock Eye movements and word skipping during reading: effects of word
  length and predictability.
\newblock {\em Journal of Experimental Psychology: Human Perception and
  Performance\/} 37(2):514.

\bibitem[{Rayner and Well(1996)}]{rayner1996}
Keith Rayner and Arnold~D Well. 1996.
\newblock Effects of contextual constraint on eye movements in reading: A
  further examination.
\newblock {\em Psychonomic Bulletin \& Review\/} 3(4):504--509.

\bibitem[{Reichle et~al.(1998)Reichle, Pollatsek, Fisher, and
  Rayner}]{reichle1998}
Erik~D Reichle, Alexander Pollatsek, Donald~L Fisher, and Keith Rayner. 1998.
\newblock Toward a model of eye movement control in reading.
\newblock {\em Psychological Review\/} 105(1):125.

\bibitem[{Reichle et~al.(2010)Reichle, Reineberg, and Schooler}]{reichle2010}
Erik~D Reichle, Andrew~E Reineberg, and Jonathan~W Schooler. 2010.
\newblock Eye movements during mindless reading.
\newblock {\em Psychological Science\/} 21(9):1300--1310.

\bibitem[{Reichle et~al.(2009)Reichle, Warren, and McConnell}]{reichle2009}
Erik~D Reichle, Tessa Warren, and Kerry McConnell. 2009.
\newblock Using ez reader to model the effects of higher level language
  processing on eye movements during reading.
\newblock {\em Psychonomic Bulletin \& Review\/} 16(1):1--21.

\bibitem[{Rello and Ballesteros(2015)}]{rello2015}
Luz Rello and Miguel Ballesteros. 2015.
\newblock Detecting readers with dyslexia using machine learning with eye
  tracking measures.
\newblock In {\em Proceedings of the 12th Web for All Conference\/}. ACM,
  page~16.

\bibitem[{Roberts and Siyanova-Chanturia(2013)}]{roberts2013}
Leah Roberts and Anna Siyanova-Chanturia. 2013.
\newblock Using eye-tracking to investigate topics in {L2} acquisition and {L2}
  processing.
\newblock {\em Studies in Second Language Acquisition\/} 35(02):213--235.

\bibitem[{Santorini(1990)}]{santorini1990}
Beatrice Santorini. 1990.
\newblock Part-of-speech tagging guidelines for the {P}enn {T}reebank project
  (3rd revision).
\newblock {\em Technical Reports (CIS)\/} .

\bibitem[{Shermis and Burstein(2013)}]{shermis2013}
Mark~D Shermis and Jill Burstein. 2013.
\newblock {\em Handbook of automated essay evaluation: Current applications and
  new directions\/}.
\newblock Routledge.

\bibitem[{Smith and Levy(2013)}]{smith2013}
Nathaniel~J Smith and Roger Levy. 2013.
\newblock The effect of word predictability on reading time is logarithmic.
\newblock {\em Cognition\/} 128(3):302--319.

\bibitem[{Stolcke et~al.(2002)}]{stolcke2002}
Andreas Stolcke et~al. 2002.
\newblock {SRILM} - an extensible language modeling toolkit.
\newblock In {\em Interspeech\/}. volume 2002, page 2002.

\bibitem[{Williamson(2009)}]{williamson2009}
David~M Williamson. 2009.
\newblock A framework for implementing automated scoring.
\newblock In {\em Annual Meeting of the American Educational Research
  Association and the National Council on Measurement in Education, San Diego,
  CA\/}.

\bibitem[{Xi et~al.(2008)Xi, Higgins, Zechner, and Williamson}]{xi2008}
Xiaoming Xi, Derrick Higgins, Klaus Zechner, and David~M Williamson. 2008.
\newblock Automated scoring of spontaneous speech using {SpeechRaterSM} v1. 0.
\newblock {\em ETS Research Report Series\/} 2008(2).

\bibitem[{Yannakoudakis et~al.(2011)Yannakoudakis, Briscoe, and
  Medlock}]{yannakoudakis2011}
Helen Yannakoudakis, Ted Briscoe, and Ben Medlock. 2011.
\newblock A new dataset and method for automatically grading {ESOL} texts.
\newblock In {\em ACL\/}. pages 180--189.

\end{thebibliography}
\bibliographystyle{acl_natbib}

%\appendix
%\section{Supplemental Material}
%\label{sec:supplemental}

\end{document}